\documentclass[runningheads]{llncs}
\usepackage[T1]{fontenc}
\usepackage{graphicx}
\usepackage{color}
\usepackage{graphicx}
\usepackage{amssymb}
\usepackage{color}
\usepackage[most]{tcolorbox}
\usepackage{pifont}
\usepackage{wrapfig}
\usepackage{tabularx}
\usepackage{hhline}
\usepackage{wrapfig}
\usepackage{cite}
\usepackage{enumitem}

\usepackage{fontawesome}

\usepackage{hyperref}
\usepackage{blindtext}
\newcommand{\dimension}[5]{
    \begin{tcolorbox}[skin=bicolor,fonttitle=\bfseries,coltitle=black,colbacktitle=#2,colback=#2!20,colframe=#2,title=#1, after skip=0.35em,left=3pt, right=3pt, top=3pt, bottom=3pt,boxsep=0pt,colbacklower=#2!10,middle=0.4em,toptitle=6pt, bottomtitle=4pt,sharp corners=all,boxrule=0mm, leftrule=1mm]
    \textbf{Opportunities}: #3
    \end{tcolorbox}%
    \vspace{1em}\noindent#5
}

\usepackage{tikz}
\usepackage{tcolorbox}
\usepackage{color,soul}

\usepackage{subcaption}
\definecolor{dataset}{HTML}{F4C2C2}
\definecolor{explanation}{HTML}{D6CADD}%D8BFD8
\definecolor{evaluation}{HTML}{BCD4E6}

\usepackage[most]{tcolorbox}

\makeatletter
\newtcbox{\kwColorBox}[1][]{on line,fontupper=\footnotesize\sffamily\bfseries\small,boxrule=1.5pt,arc=2pt,coltext=gray,colback=#1!10!white,colframe=#1,boxsep=0pt,left=1.5pt,right=1.5pt,top=1.5pt,bottom=1.5pt}
\makeatother

\newcommand{\kw}[2]{%
    \begin{kwColorBox}[#2]%
    {#1}%
    \end{kwColorBox}%
    \xspace%
}

\newcommand{\rawchallengedataset}[1]{\kw{#1}{dataset}}
\newcommand{\defD}[1]{\rawchallengedataset{#1}}

\newcommand{\rawchallengeexplanation}[1]{\kw{#1}{explanation}}
\newcommand{\defX}[1]{\rawchallengeexplanation{#1}}

\newcommand{\rawchallengeevaluation}[1]{\kw{#1}{evaluation}}
\newcommand{\defE}[1]{\rawchallengeevaluation{#1}}
\newcommand{\refRO}[1]{\hyperref[challenge:#1]{\rawchallengedonotuse{#1}}}

\begin{document}
\title{Challenges and Opportunities\\ in Text Generation Explainability}

\author{Kenza Amara\orcidID{0000-0001-7139-5562} \and
Rita Sevastjanova\orcidID{0000-0002-2629-9579} \and
Mennatallah El-Assady}
\authorrunning{K. Amara et al.}
\institute{Department of Computer Science, ETH Zurich, Switzerland\\
\email{\{kenza.amara, menna.elassady\}@ai.ethz.ch}\\
\email{rita.sevastjanova@inf.ethz.ch}}

\maketitle              % typeset the header of the contribution
\begin{abstract}
    
The necessity for interpretability in natural language processing (NLP) has risen alongside the growing prominence of large language models. Among the myriad tasks within NLP, text generation stands out as a primary objective of autoregressive models. The NLP community has begun to take a keen interest in gaining a deeper understanding of text generation, leading to the development of model-agnostic explainable artificial intelligence (xAI) methods tailored to this task. The design and evaluation of explainability methods are non-trivial since they depend on many factors involved in the text generation process, e.g., the autoregressive model and its stochastic nature. This paper outlines 17 challenges categorized into three groups that arise during the development and assessment of attribution-based explainability methods. These challenges encompass issues concerning tokenization, defining explanation similarity, determining token importance and prediction change metrics, the level of human intervention required, and the creation of suitable test datasets. The paper illustrates how these challenges can be intertwined, showcasing new opportunities for the community. These include developing probabilistic word-level explainability methods and engaging humans in the explainability pipeline, from the data design to the final evaluation, to draw robust conclusions on xAI methods.

\end{abstract}
%
%
%\section{Introduction}

\begin{figure}[t]
    \centering
    \includegraphics[width=\linewidth]{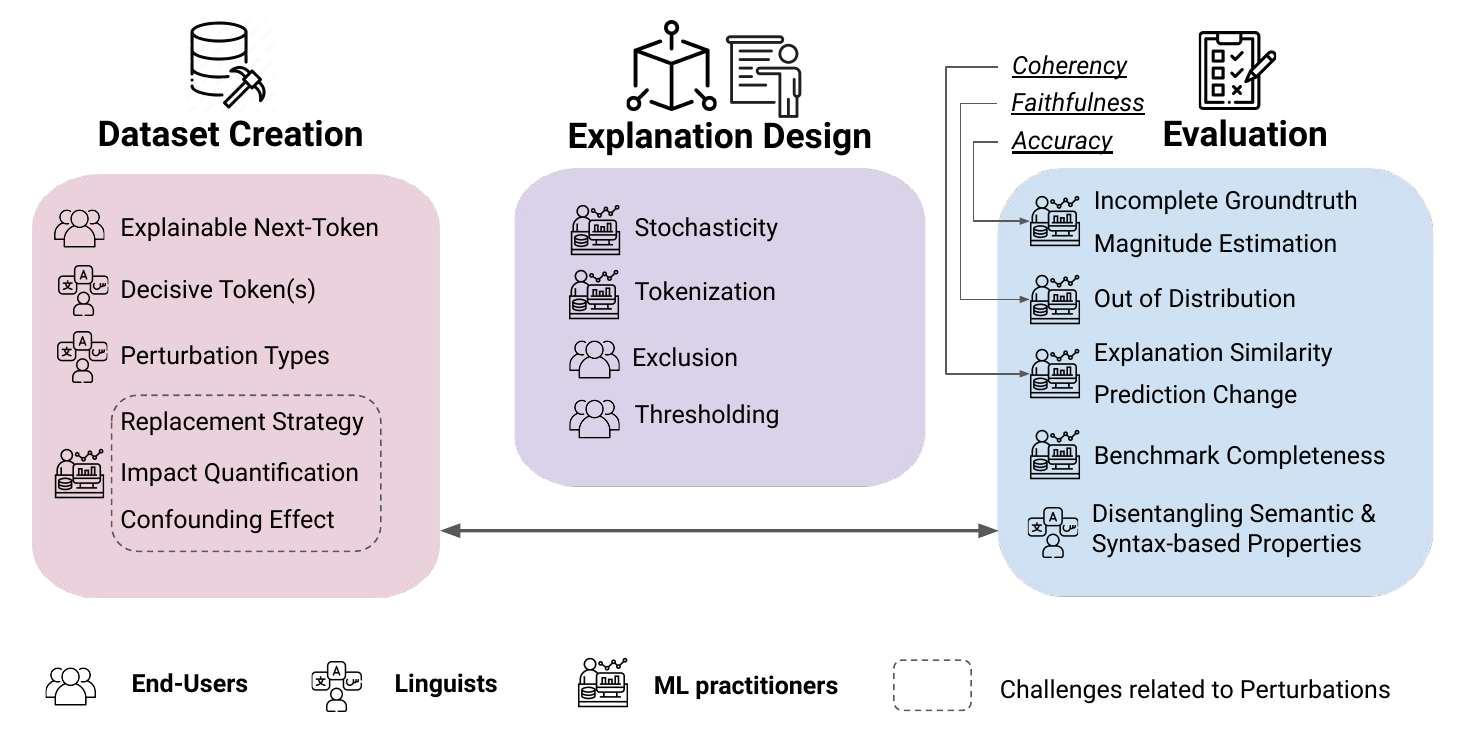}
    \caption{Challenges at various stages of the text generation explainability process, including the dataset creation, the explanation design, and the explanation evaluation. Different stakeholders address these challenges. Linguists are responsible for designing semantic, syntactic, and grammatical perturbations. End-users contribute to defining the overall format of expected explanations, e.g., size of explanations. Machine learning (ML) practitioners tackle the remaining challenges related to the language model, the data, and the evaluation metrics.}
    \label{fig:challenges_and_opportunities}
\end{figure}

\section{Introduction}

%Interpretability of language models (LMs) has gained significant importance alongside the rise of generative AI. Despite the notable achievements of recent LMs, there remains a wide range of tasks where these models struggle~\cite{truong-etal-2023-language}. Thus, it is imperative to enhance our understanding of LM reasoning and develop faithful explainability methods that align with human mental models. Consequently, there is growing interest in developing systematic evaluation frameworks that incorporate both model-centric and human-centric approaches.

Text generation tasks are ubiquitous in natural language processing (NLP), spanning a range of applications. These tasks encompass predicting the next word in a sequence, generating complete paragraphs, or even producing entire documents. Text generation can also be framed as a sequence-to-sequence task that aims to take an input sequence and generate an output sequence for machine translation and question-answering. Despite the notable achievements of recent autoregressive models to generate text, there remains a wide range of tasks where these models struggle~\cite{truong-etal-2023-language}. Thus, it is imperative to enhance our understanding of LM reasoning. However, the unique characteristics of text generation pose specific challenges regarding its explainability, e.g., challenges related to text data, autoregressive models, or tokenization. Although various strategies exist to explain the outcomes of autoregressive models—such as causal analyses of network activation subspaces, attention coefficient analysis, probing representations, and natural language representations—our paper specifically centers on attribution-based methods that provide explanatory masks. The primary rationale for this focus is that these masks can be easily transformed into importance scores for individual input features, specifically the tokens in the input sentence. This greatly enhances human interpretation of explanations and the understanding of each token's role. Furthermore, this approach enables human-based evaluation alongside model-focused evaluation, such as evaluating faithfulness.

Understanding the challenges at each stage of the explainability pipeline is crucial when explaining next token generation. During the explanation design phase, decisions are made regarding how explanatory masks should be translated into textual explanations. To ensure control over explanation evaluation, one approach is to craft datasets with specific perturbations. However, this introduces additional obstacles that machine learning (ML) practitioners must address. Moreover, making informed decisions about how metrics will be employed is imperative during the final evaluation of the explanations generated from these constructed datasets.

Addressing these challenges requires the engagement of various stakeholders at every stage of the explainability process. Currently, human involvement is limited to the final step, where they verify if explanations of the next token align with their mental models~\cite{carvalho2019machine,Guidotti2018survey}. However, research indicates that only involving humans at the end of the explainability pipeline, during user studies and quality ratings, does not provide sufficient evidence that explanations are helpful and meet human expectations~\cite{lai2019human,hase2020evaluating}. As motivated by Zhao et al.~\cite{zhao2024survey}, we propose to include different stakeholders from the start of the explainability process to gain control over the evaluation of explainable artificial intelligence (xAI) methods for text generation.

Our objective is to highlight both the challenges and opportunities inherent to explainability for text generation, encouraging the development of a comprehensive and controlled evaluation framework for explainability methods. \autoref{fig:challenges_and_opportunities} introduces the 17 challenges that emerge in dataset creation, explanation design, and method evaluation that are unique to textual data and the task of text generation. To tackle those challenges, we propose strategies that involve humans at every step of the explainability pipeline to ensure robust characterization of xAI methods. Tasks are distributed among various stakeholders, including ML practitioners, linguists, and end-users, showing the necessity of diverse knowledge sources in the explainability process. In addition, this position paper explores new opportunities in text generation explainability, such as the development of probabilistic word-level explainability methods and the design of tailored perturbations. The establishment of a benchmark comprising well-designed perturbed datasets would facilitate the generation of coherent explanations and enable controlled evaluation of xAI methods that would lead to a more nuanced understanding of their performance. To summarize our contributions, 
\begin{itemize}[label=$\circ$]
    \item We present challenges encountered at each stage of the explainability process of autoregressive LMs for text generation.
    \item We demonstrate that addressing these challenges within the explainability process involves collaborating with various stakeholders, including ML practitioners, linguists, and end-users, each contributing expertise at different stages.
    \item We introduce new opportunities for establishing a robust explainability benchmark through well-designed perturbed datasets. This initiative aims to comprehensively evaluate existing explainability methods.
\end{itemize}

Our paper serves as a roadmap for those interested in developing, evaluating, characterizing, and comparing explainability methods for text generation. Throughout the paper, the icons \defD{D}, \defX{X}, and \defE{E} are used to represent the challenges related to the \textbf{D}ataset design, the e\textbf{X}planation design, and the \textbf{E}valuation respectively.

\section{Background and Related Work}

% Check related work of CausalGym

\subsection{Text Generation}

Text generation entails initial tokenization followed by autoregressive text generation, where each token is conditioned on previously generated tokens. The tokenization phase crucially depends on the choice of tokenizer, as it establishes the language unit for the autoregressive model. Prior studies have delved into the impact of tokenization, particularly subword tokenization, on word alignment in machine translation tasks~\cite{sentencepiece,bilingualsegmentation} and its utility to make the length of parallel sentences more even~\cite{wordalignment}. While subword tokenization has been recognized in the machine translation community as a solution for handling misspellings and multilingual data, our paper demonstrates that it also introduces new challenges in text generation explainability. Moreover, the probabilistic nature of language generation techniques significantly influences model performance, user experience, and ethical considerations. Sampling methods, e.g., top-k sampling and temperature-based sampling, and exploration strategies, e.g., beam search, introduce randomness into text generation, allowing models to explore diverse linguistic possibilities and improve text quality~\cite{LMstochasticity}. However, this randomness introduces unpredictability into responses, which is not addressed in deterministic explainability procedures. To our knowledge, there has been no consideration of the impact of LM stochasticity on explainability. Thus, our paper aims to delineate the challenges posed by LM randomness in generated explanations and propose solutions to address this issue.

\subsection{Explainability Methods}

In xAI, explanations are commonly classified into several categories. Firstly, they can be categorized based on whether they pertain to an individual prediction (local) or the overall prediction process of the model (global). Secondly, explanations are distinguished based on whether they arise directly from the prediction process itself (self-explaining) or if they require additional post-processing (post-hoc)~\cite{Guidotti2018survey}. Furthermore, explanations can be classified as either pertaining to the data-level (model-agnostic) or considering the model's behavior (model-aware)~\cite{spinner2020explainer}. Regardless of their categorization, they should accurately depict the behavior of the models, i.e., be faithful) and enhance user comprehension and trust in black-box models~\cite{jacovi2020towards}. 

The importance of explainability within the NLP research is showcased by the recent survey papers. For example, Danilevsky et al.~\cite{danilevsky-etal-2020-survey} categorize methods for explaining NLP, such as feature importance and surrogate models, and outline available techniques for generating explanations for NLP model predictions. Zini and Awad~\cite{zini2022survey} examine both model-agnostic and model-specific explainability methods in NLP, categorizing them based on what they explain, including word embeddings, LM operations, and model decisions (predictions). Nagahisarchoghaei et al.~\cite{nagahisarchoghaei2023empirical} describe the importance and development of xAI research across various domains, including language modeling tasks. Sajjad et al.~\cite{Sajjad2022survey} present a survey on neuron analysis, covering methods for understanding neuron properties in NLP models, including LMs. Additionally, Vijayakumar~\cite{vijayakumar2023interpretability} focuses on the latent space (activation space), describing methods that explore neuron activations and their learning capabilities.

\subsection{Attribution-based Methods}

Recent advancements have explored model-agnostic attribution-based methods for explaining text generation tasks. While many of these methods stem from traditional feature-based approaches like SHAP~\cite{ShapleyValue} and LIME\cite{LIME}, newer methodologies have been proposed to better suit NLP tasks. In textual data, words exhibit strong interactions, and their contributions heavily depend on context. Therefore, feature attributions for textual data need to be tailored to accommodate these intricate dependencies. Existing explainability approaches for text data are predominantly tailored to classification tasks~\cite{TransShap,HierarchicalTextShapley}, with only recent attempts focusing on elucidating autoregressive models and their text generation processes. HEDGE~\cite{HierarchicalTextShapley} is an example of a SHAP-based method designed to address context dependencies specific to text data. It constructs hierarchical word clusters based on their interactions. SyntaxShap~\cite{SyntaxShap} introduces a novel approach to forming word coalitions that adhere to syntactic relationships dictated by the dependency tree, taking into account the syntactic dependencies fundamental to linguistics. Our research aims to highlight the myriad challenges associated with the project of explaining text generation with model-agnostic attribution-based methods and provide guidelines on how to address these novel issues effectively.

\section{Dataset Creation}\label{sec:dataset_creation}

\begin{figure}
    \centering
    \includegraphics[width=\linewidth]{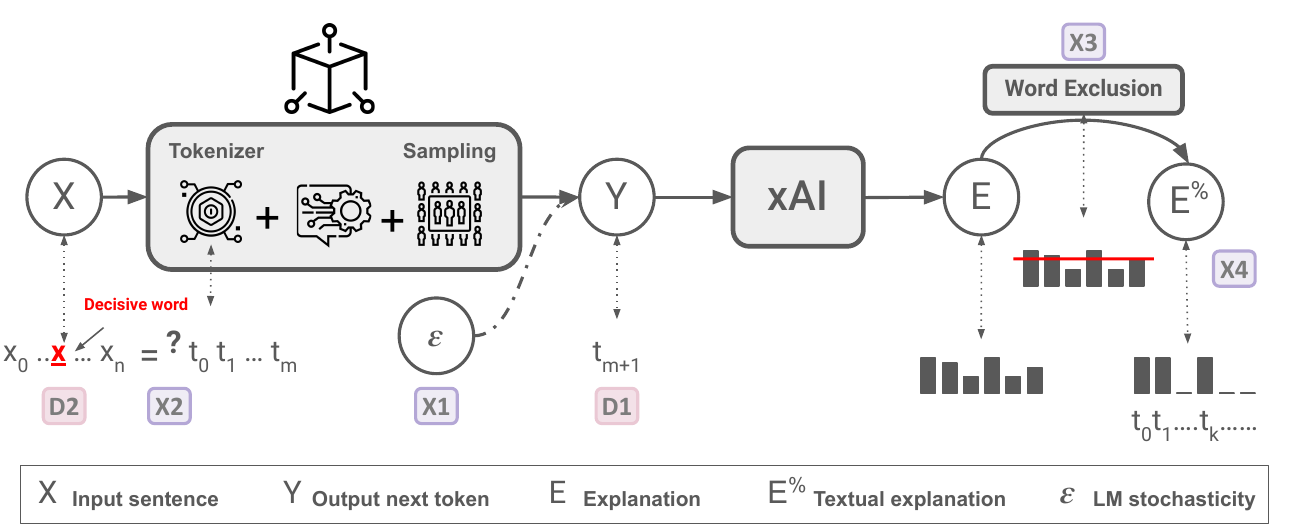}
    \caption{Example of an explanation pipeline for text generation. Challenges related to the dataset creation are referred to by icon \defD{D}, and the ones related to the explanation design with icon \defX{X}. The autoregressive language model that includes the tokenizer, the decoder, and the sampling method, takes as input an unfinished sentence $X$ and generates the next token $Y$. The explainability method then produces an explanation $E$ that is transformed into a textual explanation $E^{\%}$ after a step of thresholding and word exclusion.}
    \label{fig:perturb_explain}
\end{figure}

To achieve thorough explainability, it is essential to incorporate human-centered evaluation for constructing reliable and trustworthy models. However, involving humans at the very final stages of the xAI pipeline reveals numerous limitations. As already motivated by Zhao et al.~\cite{zhao2024survey}, we advocate for early human involvement in designing tailored datasets. This approach presents several challenges, which are elaborated in this section and referred to with icons \defD{D} in \autoref{fig:perturb_explain} and \autoref{fig:perturb_evaluate_characterize}.

\subsection{Human-centered Explanations}\label{human-centered-X}

In text generation, an explanation is represented as a vector of importance scores assigned to each token within the input sentence. These scores represent the contribution of individual tokens to the language model's prediction of the next token. We need to construct input/output pairs, which are expected to be both necessary and engaging for humans. In other words, the output should merit explanation, and there should be identifiable explanatory tokens within the input.\\

\noindent\defD{D1} \textbf{Explainable Next-Token} An explanation in text generation is crafted to explain a specific next-token, also referred to as a target. It falls upon humans to designate which tokens are worthy of explanation. A next token is deemed \textit{explainable} if there exist tokens within the input sentences that significantly influence its prediction~\cite{explanationprediction}. For example, function words like determinants are often not considered explainable since they lack semantic or grammatical dependency on preceding tokens. 
\begin{align*}
    & \text{My grandmother is \textbf{a}}\\
    & \text{My grandmother is a \textbf{loving}}
\end{align*}

\dimension{}{dataset}{Strategies to ensure that targets in text generation are explainable could be to bring end-users to participate in dataset creation. This approach enables the evaluation of xAI methods on sentences where preceding tokens strongly influence the next token.}

\noindent\defD{D2}\textbf{Decisive Words} Selecting the decisive token(s) for perturbation poses a significant challenge. Prior studies have introduced the concept of a direct causal effect of a token on prediction by examining which token, when altered, induces the most substantial change in prediction~\cite{causalgym,syntaxgym}. The tokens chosen for perturbation should profoundly influence prediction by being both necessary and sufficient for it~\cite{necessitysufficiency}. These tokens typically exhibit semantic richness, such as content words (nouns) or words containing crucial information, e.g., age, gender.
\begin{align*}
    & \text{My grandmother is a \underline{loving} individual, however she also tends to be}\\
   & \text{My grandmother is a \underline{strict} individual, however she also tends to be}
\end{align*}

\dimension{}{dataset}{Identifying the decisive tokens requires the intervention of linguists, although automation using a well-trained neural network could be considered.}

\subsection{Perturbation-based Datasets}

Utilizing automatic or human-controlled perturbations on textual inputs has proven effective in assessing the robustness of NLP systems \cite{Moradi2021EvaluatingTR}. Perturbation methods have been used for text translation \cite{niu-etal-2020-evaluating}, bias analysis \cite{prabhakaran-etal-2019-perturbation}, robustness to adversarial atacks \cite{Alshemali2020ImprovingTR}, etc. Also, when it comes to text generation task, to control the evaluation of xAI methods based on prior knowledge of decisive tokens influencing explainable next-tokens, it is advisable to construct perturbed datasets where these key tokens are identified. However, this design process presents several challenges, including managing the wide variety of perturbation types and determining how to apply these perturbations to text data effectively.\\

\begin{figure}
    \centering
    \includegraphics[width=\linewidth]{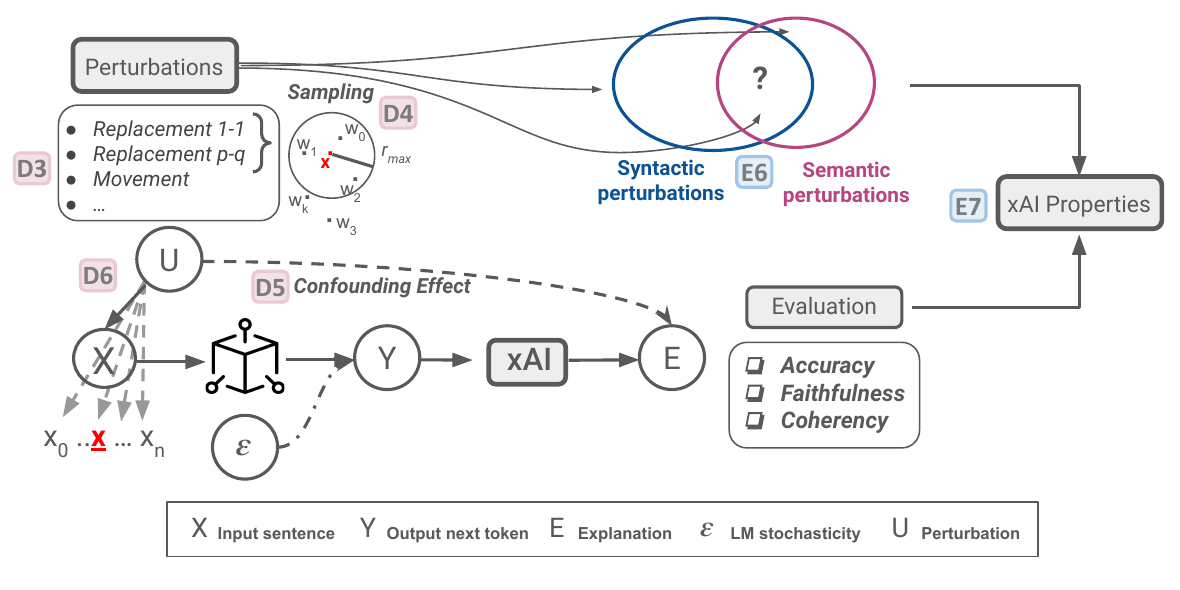}
    \caption{Evaluation procedure with well-designed perturbations. Different types of perturbations are used to create tailored datasets. The selected perturbation $U$ affects the input sentence $X$ and the explanation $E$ generated by the xAI method. By assigning linguistic properties to the different types of perturbation, it is possible to use the results of the evaluation phase to characterize xAI methods. Icons \defD{D} describe challenges related to the dataset creation and icons \defE{E} describe challenges in the evaluation phase of explanations.}
    \label{fig:perturb_evaluate_characterize}
\end{figure}

\noindent\defD{D3}\textbf{Diverse Perturbation Strategies} A perturbation consists of modifying an input sentence with a potential impact on the next token and/or the explanation. There are many ways how to perturb a sentence: 1-to-1 token replacement, multiple tokens replacement, token movement, formulation change, gender/number modification~\cite{perturbationsurvey}. Humans play a role in creating perturbations by selecting input/output pairs that produce explainable outputs and include decisive tokens. These perturbations can take various forms, such as syntactic, semantic, or grammatical. When designing perturbations, humans ensure their validity, acting as safeguards. While grammatical perturbations can be automated, generating semantic perturbations is more complex and less straightforward. By design, we expect the perturbed word(s) to be decisive and, therefore, to be of high importance in the explanation. The choice of the perturbation type will determine the number of decisive tokens to be altered. We encourage intervention on single tokens to facilitate the study and be certain to identify the full causal effect of the perturbation. Working with pairs of sentences that differ by one perturbation, we also expect their explanations to have some similarity or dissimilarity, based on the perturbation strategy chosen. 

\dimension{}{dataset}{Well-designed perturbed datasets by linguists will help rigorously evaluate the accuracy and coherency of explainability methods. As far as we know, there has not been a comprehensive benchmark of perturbed datasets to assess and categorize explainability methods for text generation. This presents new opportunities for creating well-designed perturbed datasets to evaluate xAI methods.}

\noindent\defD{D4}\textbf{Sampling Strategy to Replace Tokens} One type of perturbation is to replace a token with another one that has either a close or very different meaning while keeping the same function~\cite{wordreplacement}. A strategy for selecting a replacement word is to choose a word from a subset of words in the embedding space, where the distance between data points indicates the similarity between words. However, selecting the appropriate vector-embedding model is a non-trivial task, and it remains uncertain which layers in Language Models (LMs) effectively capture semantic information~\cite{lmfingerprints}. For minor semantic perturbations, a maximum distance $r_{max}$ is initially determined, and a word at a distance $d$ from the original word is selected such that $d<r_{max}$. Respectively, for large perturbations, a minimum distance $r_{min}$ is initially chosen, and a word at a distance $d$ from the original word is selected such that $d>r_{min}$. The selection of the minimum or maximum radius is subjective and should be justified prior to sampling words for perturbation-based analysis.
\begin{align*}
    & \text{My sister went to the \underline{park} and saw a }\\
   & \text{My sister went to the \underline{school} and saw a}\\
      & \text{My sister went to the \underline{basement} and saw a}
\end{align*}

\dimension{}{dataset}{The method of token replacement remains an unresolved question, and there are numerous avenues for ML practitioners to explore in proposing new intelligent replacement strategies and finding expressive embedding spaces.}

\subsection{Tracing the Effect of Perturbations}
%Figure with causal tree U perturbation --> Y Model'prediction --> X Explanation 

Creating perturbations is a method to acquire comprehensive insight into expected explanations. Nevertheless, there could be hidden effects of the designed perturbations. Grasping how perturbations spread helps measure the true impact of interventions on the explanations.\\
 
\noindent\defD{D5}\textbf{Indirect and Direct Effect of Input Pertubations} Perturbations on the input are designed to influence directly the prediction and, consequently, the explanation. This is the indirect impact of the perturbation on the explanation. However, if the perturbation implies changing the form of the input, e.g., number of tokens or position of tokens, the perturbation also has some direct impact on the explanation. Disentangling the direct and indirect effects of perturbations on explanations is a major challenge in perturbation-based analysis for xAI methods.

\dimension{}{dataset}{ML practitioners must explore the multiple effects of perturbations on the different variables in the explainability process. Introduce a causal graph to represent the effect of perturbations on the different variables in our problem might help grasp better their real effect. }\\

\noindent\defD{D6}\textbf{Measuring the Impact of Perturbations} A perturbation is made on one or several tokens. Explanations are expected to reflect the importance of those perturbed tokens. To measure how much a perturbation in the input sentence affects the explanation, the
discrete nature of natural language makes it more
challenging~\cite{perturbscore}. One can be interested in the attribution values of the (i) perturbed tokens or (ii) the whole sentence. In case (i), one can report the changes in the importance score of the decisive tokens; while in case (ii), one can average those changes over all words in the sentence or compute the divergence between the initial attribution distribution and the new one obtained with the perturbed sentence. 

\dimension{}{dataset}{There are many approaches to measuring the impact of perturbations on explanations that ML practitioners can explore in the future.}

\section{Explanation Design}

Explanations for text generation can take many forms. In this section, we explore the challenges that arise when generating and transforming explanations for the final explainability evaluation. Those originate from the language model or the nature of the data and are presented in \autoref{fig:perturb_explain} with icons \defX{X}.

\subsection{Challenges Originating from the Language Model}

Methods for explainability produce explanations that are closely tied to the specific model being used. Consequently, any limitations or biases present in the model will directly affect the explanations provided.\\ 

\noindent\defX{X1}\textbf{Stochasticity} Explainability methods involve deterministic computations of how each word contributes to the model's prediction of the next token~\cite{HierarchicalTextShapley,SyntaxShap}. For a given next token, the explanation varies uniquely for each algorithm (such as SHAP~\cite{ShapleyValue}, LIME~\cite{LIME}, etc.). However, altering the random seed can affect the outcome after sampling a prediction from the probability distribution generated by the autoregressive language model, and consequently altering the entire explanation~\cite{LMstochasticity}. The challenge lies in reconciling the sampling method's inherent probabilistic nature with the uniqueness of explanations.

\dimension{}{explanation}{We encourage ML practitioners to develop explanations that can withstand the inherent stochasticity of language models and their sampling strategy. One direction could be to develop probabilistic explainability methods that do not merely replicate randomly sampled final outputs but instead construct (approximate) explanations that reflect the probability distribution across the entire (top-k) vocabulary. Moreover, the inherent probabilistic nature of language models could be leveraged as a means to evaluate the robustness of explainability methods concerning variations in predictions.}

\noindent\defX{X2}\textbf{Tokenization} Explanations involve attributing value to individual words within input sentences. However, in NLP tasks, the tokenizer vocabulary may not align perfectly with our vocabulary of words, leading to situations where words are divided into multiple tokens. The approach of representing words as sequences of subword units is founded on the concept that various word categories can be conveyed using smaller units compared to whole words~\cite{sennrich-etal-2016-neural}. Subword tokenization not only allows the model to sustain a manageable vocabulary size while obtaining significant context-independent representations but also equips the model to address unfamiliar words by decomposing them into recognizable subwords~\cite{acs-etal-2021-subword}. This introduces complexity in assigning importance to features. It raises questions about how tokens stemming from the same word should be treated in terms of importance. The following example shows how subword tokenization can affect the attribution of xAI methods, making it difficult to interpret explanations:
\begin{align*}
    &\text{\textbf{Words:}\hspace{1em}}  \text{My grandpa went to Himalayas and saw a}\\
    &\hspace{1.5em}	\Downarrow\\
    &\text{\textbf{Tokens:}\hspace{1em}} \text{My $\vert$ grand $\vert$ pa $\vert$ went $\vert$ to $\vert$ Himal $\vert$ ayas $\vert$ and $\vert$ saw $\vert$ a}
\end{align*}

% \begin{align*}
%     & \text{My $\vert$ sister $\vert$ went $\vert$ to $\vert$ Himal $\vert$ ayas $\vert$ and $\vert$ saw $\vert$ a}\\
%     & \text{My $\vert$ grand $\vert$ pa $\vert$ always $\vert$ brings $\vert$ a}
% \end{align*}

\dimension{}{explanation}{ML practitioners could explore strategies for handling tokens originating from the same word during explanation evaluation, such as assigning them the average importance. Looking ahead, we encourage ML practitioners to develop explainability methods that can effectively address the subword tokenization problem. For instance, in shap-based xAI methods, we could consider tokens belonging to the same word as a single player in the game.}

\subsection{Challenges Originating from the Text Data}\label{sec:textual_x}

After the explainability methods assign importance scores to each token in the input sentence, translating this into a human-intelligible format presents further challenges.\\

\noindent\defX{X3}\textbf{Exclusion} When dealing with text data, a significant challenge arises in the process of converting an explanatory mask – a vector of importance scores – into a textual explanation (so called \textit{predicted rationales}\cite{deyoung-etal-2020-eraser}). This process involves transforming explanatory masks into binary vectors by applying a chosen threshold, where words are retained or excluded based on whether the vector value is 1 or 0 at their respective positions. The following example illustrates the word exclusion problem. The intensity of orange indicates the magnitude of importance scores.
\begin{equation*}
    \begin{cases}
        &\text{\colorbox{orange!10!white}{After} \colorbox{orange!20!white}{dinner}, \colorbox{orange!5!white}{they} \colorbox{orange!10!white}{are} \colorbox{orange!90!white}{not}}\\
        &\text{threshold = \colorbox{orange!20!white}{  X }}
    \end{cases}
\Bigg\} \rightarrow \text{...\colorbox{orange!20!white}{dinner}, ... \colorbox{orange!90!white}{not}}
\end{equation*}

However, the conversion of scalar importance values into binary form results in a substantial loss of information, hindering the ability to compare the importance of words effectively. 

\dimension{}{explanation}{There is space for end-users to explore new replacement strategy such as a weighted-word replacement. Rather than relying on a threshold to determine whether to retain or remove a word, this approach involves replacing words with others that reflect their importance scores through similarity. Specifically, if a word $x$ in the input sentence is assigned a weight $w$, it is replaced with a word at a distance proportional to $1/w$. Consequently, a word with an importance score of $0$ would be replaced by a random word, offering a more nuanced and informative approach to textual explanations. With shared knowledge, end-users can approximate which words are $w$-similar in the replacement strategy.}

\noindent\defX{X4}\textbf{Thresholding} Explanations are generated as a weighted vector on the input tokens. They are converted into textual explanations as described above. This requires choosing a certain threshold to decide which token can be considered as important. Selecting the threshold for sparsing the explanation relies on subjective human judgment, introducing variability. The number of tokens to keep is highly dependent on the context and next generated token and therefore requires human involvement.

\dimension{}{explanation}{With human prior knowledge or intuition, the explanations can be reduced by end-users to the top-ranked words. In some situations, only one token in the input sentence is decisive for the next token, while in some other cases, multiple tokens carry useful information for the following text.}

\section{Explanation Evaluation}\label{sec:evaluation}

\begin{figure}
    \centering
    \includegraphics[width=\linewidth]{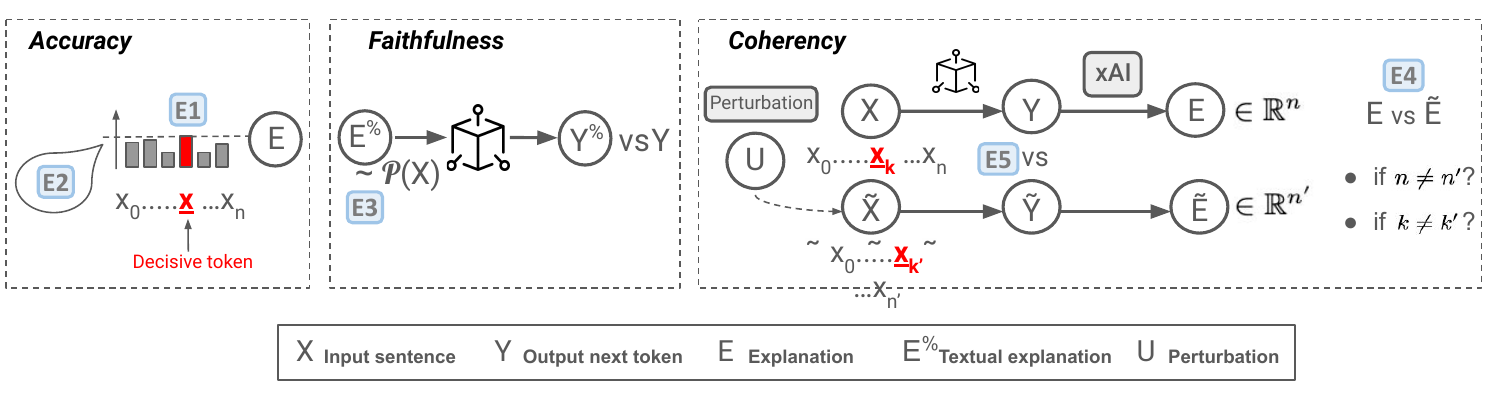}
    \caption{Quantitative evaluation of explainability employs three key quality metrics. While accuracy and faithfulness assess explanations at the instance level, coherency adopts a contrastive approach, evaluating pairs of sentences, including an input sentence and its perturbed counterpart. Challenges are inherent to each metric and are represented by the icon \defE{E}.}
    \label{fig:evaluation}
\end{figure}

Evaluating explanations for well-designed perturbed datasets with current xAI metrics presents new challenges, notably the ongoing debate on defining adequacy, estimating explanation similarity, and evaluating with incomplete prior knowledge. In addition, these metrics carry inherent limitations. However, the main challenges remain in interpreting evaluation results to refine method characterization, with the persistent question of feasibility. All challenges related to xAI evaluation are presented in \autoref{fig:perturb_evaluate_characterize} and \autoref{fig:evaluation} with icons \defE{E}.

\subsection{Static Evaluation with Accuracy}

Once an explanation has been generated for explaining a specific prediction, i.e., next token to the input sentence, one can measure its similarity to some groundtruth explanation. In the field of xAI, getting access to this prior knowledge is difficult and we often rely on some domain experts' intuition or scientific knowledge. However, by constructing a well-designed perturbation benchmark, we can now confidently use accuracy since we have clear expectations of what should be the explanations.\\

\noindent\defE{E1}\textbf{Limited Groundtruth Explanation} When ground truth explanations are accessible, accurate evaluation becomes feasible. For instance, for the classification task, there exist benchmarks such as ERASER \cite{deyoung-etal-2020-eraser} that presents a collection of human annotated \textit{rationales} used as ground truth for explanation evaluation. In the scenario of well-constructed datasets, where the next token's predictability is clear and the decisive tokens in the input sentence are known, these decisive tokens are expected to hold significant importance, serving as the ground truth explanation~\cite{syntheticgroundtruth}. However, determining the expected importance scores for the remaining tokens in the input sentence poses a challenge, as our understanding is confined to the decisive tokens, which implies disregarding the rest. 

\dimension{}{evaluation}{Given our incomplete knowledge of the ground truth explanations, there is a need to develop new strategies for integrating the remaining tokens, namely the non-decisive ones, and establishing comprehensive ground truth for the entire input sentence. One approach ML practitioners could adopt, involves conducting a separate comparison between the decisive tokens and the remaining ones.}

\noindent\defE{E2}\textbf{Magnitude Estimation} A recurrent obstacle in machine learning evaluation is determining the threshold for what constitutes satisfactory performance~\cite{goodenoughlanguage,goodenough}. In other words, one must address the question: what is good enough? One approach to measuring accuracy is ensuring that the decisive tokens rank highly within the explanation. Alternatively, one can leverage prior knowledge of decisive tokens to create a binary vector and compute cosine similarity with the explanatory vector. Confirming that decisive tokens receive an importance score above a predetermined threshold is another option.

\dimension{}{evaluation}{Faced with multiple options to estimate accuracy, it is imperative to justify the chosen method. Furthermore, ML practitioners must establish a more rigorous and systematic accuracy evaluation procedure, particularly when dealing with partial ground truth.}

\subsection{Static Evaluation with Faithfulness}

Explanations are also evaluated with model-based evaluation metrics such as the popular faithfulness or fidelity metric~\cite{jacovi-goldberg-2020-towards}. It evaluates the capacity of the model to retrieve the initial prediction. The explanation is given as input to the model. The new prediction should be close to the initial one if the explanation is faithful to the model. However, this metric is only meaningful if the prediction is explainable (see section~\ref{human-centered-X}). If the generated next token is not chosen to be explainable, any explanations can be faithful and none will be very informative for the users.\\

\noindent\defE{E3}\textbf{Out-of-distribution Problem} Faithfulness comes with common limitations such as the out-of-distribution problem well-known in the field of explainability working with image data~\cite{OOD-pb} or graph data~\cite{OOD-GNN}. Based on a ``removal'' strategy, i.e., we remove or keep the entities estimated as important, faithfulness withdraws some entities by setting them to a baseline value either replacing them with the PAD token or with a random token from the tokenizer vocabulary. \cite{Hsieh_undated-vh} correctly observe that this evaluation procedure favors words that are far away from the baseline, i.e., semantically dissimilar. Consequently, methods that focus on tokens distant from the PAD baseline are favored. In addition, truncated sentences after token removal can lie out of the data distribution used for training the language model. In this case, model behavior might differ not because of removing important information but because of evaluating a sample outside the training distribution.

\dimension{}{evaluation}{To address this problem, the role of ML practitioners is to verify that generated explanations are not OOD or they should adopt a re-training strategy~\cite{GinX-Eval,Hooker2018-vg}.}

\subsection{Contrastive Evaluation with Coherency}

Evaluating explanations for the next token generated in datasets featuring sentence-paired perturbations can be conducted using coherency as a metric~\cite{SyntaxShap}. This evaluative approach, known as contrastive evaluation, involves comparing the explanations of two sentences within a pair, which differ due to perturbed tokens, against prior expectations derived from the perturbation's nature. However, this comparison is not trivial, as perturbations may also impact the structure of the explanations themselves, adding a layer of complexity to the evaluation process~\cite{contrastivexai}.\\

\noindent\defE{E4}\textbf{Explanation Similarity Estimation} In perturbation-based analysis, it is possible to evaluate explanations with respect to their perturbed counterpart (e.g., through \textit{counterfactual reasoning}~\cite{wu-etal-2021-polyjuice}). These sentence-paired explanations are expected to have a certain degree of similarity, which is often correlated with the similarity of the paired predictions. For instance, given a pair of negative sentences which differs by the position and form of the negation, we expect the prediction to be identical. The explanations are expected to give the same importance to the negations, i.e., the decisive tokens, and the rest, like in the following example where the prediction is \textbf{hungry} and the importance of input tokens highlighted in orange:
\begin{align*}
    &\text{\colorbox{orange!10!white}{After} \colorbox{orange!20!white}{dinner}, \colorbox{orange!5!white}{they} \colorbox{orange!10!white}{are} \colorbox{orange!90!white}{not} \textbf{hungry}.}\\
    &\text{\colorbox{orange!10!white}{After} \colorbox{orange!20!white}{dinner}, \colorbox{orange!80!white}{none} \colorbox{orange!0!white}{of} \colorbox{orange!5!white}{them} \colorbox{orange!10!white}{is} \textbf{hungry}.}
\end{align*}
However, comparing explanations becomes less straightforward when perturbations alter the sentence length or the positions of input tokens~\cite{chunkalignment}. Thus, contrastive evaluation presents a new challenge: defining an appropriate strategy for computing explanation similarity~\cite{similaritychallenge}.

\dimension{}{evaluation}{ML practitioners must find different ways of computing explanations similarity for each perturbation type evaluated with the coherency metric. Considering the nature of the perturbation, they must determine a threshold of proximity to assert explanation similarity and propose a method for comparing explanations of varying lengths. One possible approach could involve separating the perturbed tokens from the rest and computing the similarity of these two segments separately.}

\noindent\defE{E5}\textbf{Estimation of Prediction Change} Certain perturbations are expected to induce substantial changes in the model's predictions. This is the case for semantic perturbations, where substituting the decisive token with a semantically distinct one is expected to have a pronounced effect on the prediction. On the other hand, syntactic perturbations are expected to induce little change in the prediction~\cite{syntaxperturbation}. Nonetheless, determining the threshold for what constitutes a significant or a minor change in prediction is a non-trivial task~\cite{goodenough,goodenoughlanguage}.

\dimension{}{evaluation}{In the dataset creation phase, ML practitioners with the help of linguists and the feedback of end-users, must select perturbations that unquestionably induce changes in predictions, such as altering the verb from singular to plural form in accordance with the subject; or perturbations for which it is undeniable that the prediction semantic meaning has not change.}

\subsection{Characterization of Explainability Methods}

After evaluating explainability methods on purpose-built datasets (see section~\ref{sec:dataset_creation}), both end-users and ML practitioners are keen on characterizing these methods to discern their appropriate applications. Beginning with an analysis of the perturbation's nature, methods can be characterized based on their robustness against specific syntactic or semantic alterations. This characterization process is further refined by considering the perturbation strategy, such as replacement, movement, or constituent change. However, this characterization step is not without its complexities and poses two major challenges.\\

\noindent\defE{E6}\textbf{Disentangling Syntactic and Semantic Perturbations} By using well-defined perturbations, it is possible to identify interesting properties of explainability methods such as their sensitivity to syntax or semantic changes. Semantic-only perturbations do not change the syntax, e.g., replacing a token with a semantically different one that has the same role in the sentence. However, syntax-only perturbations are usually difficult to design without influencing the semantics of the sentence~\cite{costa2008adverbs,hussein2018homogeneous}. While certain syntax perturbations, like relocating adverbs or interchanging adjectives for the same noun, might maintain semantic integrity, substitutions of constituents or utilization of equivalents may induce semantic alterations. For instance, in the following example, replacing the constituent \underline{their friends} alters the meaning of the sentence:
\begin{align*}
    &\text{The students saw  \underline{their friends} after class.}\\
    &\text{The students saw  \underline{them} after class.}
\end{align*}

\dimension{}{evaluation}{In order to thoroughly characterize explainability methods, it is crucial to separate linguistics properties such as robustness to syntactic or semantic variations. This distinction should be already inherent to the nature of the perturbations and therefore accounted for in the dataset creation. This is the role of linguists to propose minimal perturbations along a single dimension to regulate changes in predictions and explanations~\cite{disentanglesynsem}, thereby establishing clear criteria for a rigorous evaluation.}

\noindent\defE{E7}\textbf{Benchmark Completeness} The vast number of potential perturbations in syntax, semantics, grammar, and other linguistic elements renders it impractical to assess explainability across all variations. Moreover, it remains uncertain whether such a comprehensive benchmark exists and could cover all crucial properties necessary for fully defining xAI methods in text generation. Diverse perturbed datasets could however shed light on characteristics of the methods and their linguistic robustness~\cite{perturbrobustness,perturbationsurvey} inspired by behavioral and structural probing~\cite{posthocxAI}, and help users in choosing appropriate methods for their problem.

\dimension{}{evaluation}{Many opportunities exist for ML practitioners to start identifying directions to better characterize explainability methods. This initial step could pave the way for crafting a taxonomy delineating the main properties of explainability methods, which can be unveiled through datasets tailored for this purpose.}

\section{Conclusion}

\begin{figure}[h]
    \centering
    \includegraphics[width=\linewidth]{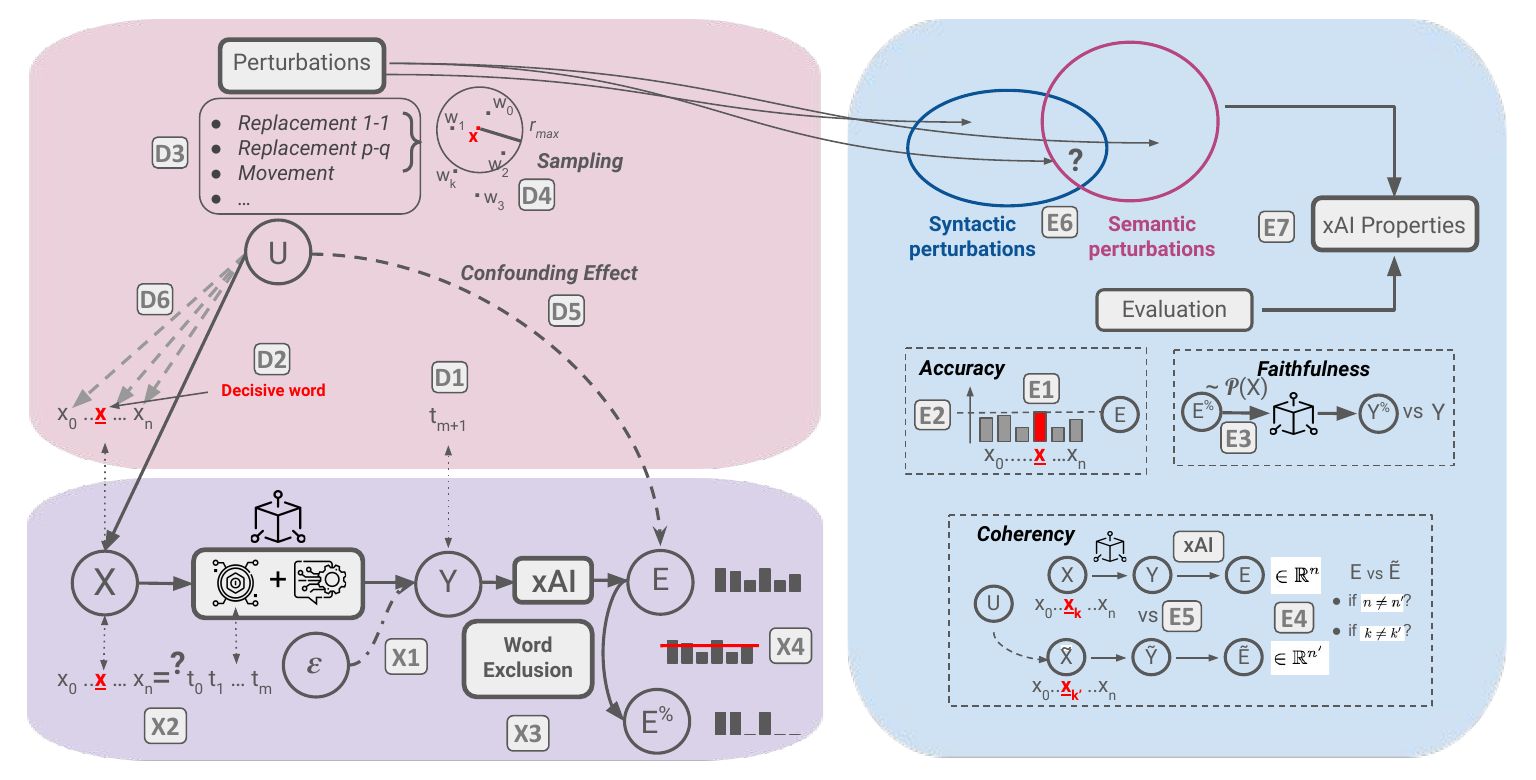}
    \caption{Summary of the challenges in explainability for text generation. Challenges arise in the \textcolor{dataset!200}{\textbf{dataset creation}}, the \textcolor{explanation!200}{\textbf{explanation design}} and the \textcolor{evaluation!200}{\textbf{evaluation}} to rigorously characterize explainability methods.}
    \label{fig:summary}
\end{figure}

Explaining next-token generation is a growing area of research. In this position paper, we demonstrate that challenges emerge at each stage of the explanation process for text generation, encompassing evaluation and perturbation-based analysis. Our focus was on the English language; however, all the identified challenges are applicable to any other language except for data-related challenges \defD{D1}, \defD{D2}, \defD{D3}, and \defD{D4} and the tokenization challenge \defX{X2}. The challenges in this paper summarized in \autoref{fig:summary} often necessitate human intervention. Moreover, they present many opportunities to propose more robust xAI evaluation through well-designed perturbed datasets. By establishing a standardized benchmark of tailored datasets with specific perturbations, we can identify important properties of existing explainability methods and conduct thorough comparisons. Currently, there is no comprehensive benchmark to rigorously characterize xAI methods for text generation. Thus, this paper also serves as a call to action for the creation of a comprehensive benchmark to evaluate xAI methods for text generation across various dimensions, including semantic comprehension, syntactic robustness, and grammatical fidelity.

\begin{credits}
\subsubsection{\discintname}
The authors have no competing interests to declare that are relevant to the content of this article.
\end{credits}

\newpage
\bibliographystyle{splncs04}
\bibliography{mybibliography}

\begin{thebibliography}{10}
\providecommand{\url}[1]{\texttt{#1}}
\providecommand{\urlprefix}{URL }
\providecommand{\doi}[1]{https://doi.org/#1}

\bibitem{acs-etal-2021-subword}
{\'A}cs, J., K{\'a}d{\'a}r, {\'A}., Kornai, A.: Subword pooling makes a difference. In: Proc. of the 16th Conf. of the European Chapter of the Association for Computational Linguistics: Main Volume. pp. 2284--2295. Association for Computational Linguistics, Online (Apr 2021). \doi{10.18653/v1/2021.eacl-main.194}, \url{https://aclanthology.org/2021.eacl-main.194}

\bibitem{Alshemali2020ImprovingTR}
Alshemali, B., Kalita, J.K.: Improving the reliability of deep neural networks in nlp: A review. Knowl. Based Syst.  \textbf{191},  105210 (2020), \url{https://api.semanticscholar.org/CorpusID:209100013}

\bibitem{GinX-Eval}
Amara, K., El-Assady, M., Ying, R.: Ginx-eval: Towards in-distribution evaluation of graph neural network explanations. arXiv preprint arXiv:2309.16223  (2023)

\bibitem{SyntaxShap}
Amara, K., Sevastjanova, R., El-Assady, M.: Syntaxshap: Syntax-aware explainability method for text generation. arXiv preprint arXiv:2402.09259  (2024)

\bibitem{causalgym}
Arora, A., Jurafsky, D., Potts, C.: Causalgym: Benchmarking causal interpretability methods on linguistic tasks. arXiv preprint arXiv:2402.12560  (2024)

\bibitem{carvalho2019machine}
Carvalho, D.V., Pereira, E.M., Cardoso, J.S.: Machine learning interpretability: A survey on methods and metrics. Electronics  \textbf{8}(8), ~832 (2019)

\bibitem{HierarchicalTextShapley}
Chen, H., Zheng, G., Ji, Y.: Generating hierarchical explanations on text classification via feature interaction detection. In: Jurafsky, D., Chai, J., Schluter, N., Tetreault, J. (eds.) Proceedings of the 58th Annual Meeting of the Association for Computational Linguistics. pp. 5578--5593. Association for Computational Linguistics, Online (Jul 2020). \doi{10.18653/v1/2020.acl-main.494}, \url{https://aclanthology.org/2020.acl-main.494}

\bibitem{costa2008adverbs}
Costa, J.: Adverbs and the syntax-semantics interplay. Estudos Lingu{\'\i}sticos  \textbf{2},  13--25 (2008)

\bibitem{danilevsky-etal-2020-survey}
Danilevsky, M., Qian, K., Aharonov, R., Katsis, Y., Kawas, B., Sen, P.: A survey of the state of explainable {AI} for natural language processing. In: Proc. of the 1st Conf. of the Asia-Pacific Chapter of the Association for Computational Linguistics and the 10th Int. Joint Conf. on Natural Language Processing. pp. 447--459. Association for Computational Linguistics, Suzhou, China (Dec 2020), \url{https://aclanthology.org/2020.aacl-main.46}

\bibitem{bilingualsegmentation}
Deguchi, H., Utiyama, M., Tamura, A., Ninomiya, T., Sumita, E.: Bilingual subword segmentation for neural machine translation. In: Proceedings of the 28th International Conference on Computational Linguistics. pp. 4287--4297 (2020)

\bibitem{similaritychallenge}
Deshpande, A., Jimenez, C.E., Chen, H., Murahari, V., Graf, V., Rajpurohit, T., Kalyan, A., Chen, D., Narasimhan, K.: Csts: Conditional semantic textual similarity. arXiv preprint arXiv:2305.15093  (2023)

\bibitem{deyoung-etal-2020-eraser}
DeYoung, J., Jain, S., Rajani, N.F., Lehman, E., Xiong, C., Socher, R., Wallace, B.C.: {ERASER}: {A} benchmark to evaluate rationalized {NLP} models. In: Jurafsky, D., Chai, J., Schluter, N., Tetreault, J. (eds.) Proceedings of the 58th Annual Meeting of the Association for Computational Linguistics. pp. 4443--4458. Association for Computational Linguistics, Online (Jul 2020). \doi{10.18653/v1/2020.acl-main.408}, \url{https://aclanthology.org/2020.acl-main.408}

\bibitem{goodenoughlanguage}
Ferreira, F., Engelhardt, P.E., Jones, M.W.: Good enough language processing: A satisficing approach. In: Proceedings of the 31st Annual conference of the Cognitive Science Society. vol.~1, pp. 413--418. Cognitive Science Society Austin, TX (2009)

\bibitem{syntaxgym}
Gauthier, J., Hu, J., Wilcox, E., Qian, P., Levy, R.: {S}yntax{G}ym: An online platform for targeted evaluation of language models. In: Celikyilmaz, A., Wen, T.H. (eds.) Proceedings of the 58th Annual Meeting of the Association for Computational Linguistics: System Demonstrations. pp. 70--76. Association for Computational Linguistics, Online (Jul 2020). \doi{10.18653/v1/2020.acl-demos.10}, \url{https://aclanthology.org/2020.acl-demos.10}

\bibitem{syntheticgroundtruth}
Guidotti, R.: Evaluating local explanation methods on ground truth. Artificial Intelligence  \textbf{291},  103428 (2021)

\bibitem{Guidotti2018survey}
Guidotti, R., Monreale, A., Ruggieri, S., Turini, F., Giannotti, F., Pedreschi, D.: A survey of methods for explaining black box models. ACM Comput. Surv.  \textbf{51}(5) (aug 2018). \doi{10.1145/3236009}, \url{https://doi.org/10.1145/3236009}

\bibitem{hase2020evaluating}
Hase, P., Bansal, M.: Evaluating explainable ai: Which algorithmic explanations help users predict model behavior? arXiv preprint arXiv:2005.01831  (2020)

\bibitem{wordreplacement}
Hiraoka, T., Takase, S., Uchiumi, K., Keyaki, A., Okazaki, N.: Word-level perturbation considering word length and compositional subwords. In: Findings of the Association for Computational Linguistics: ACL 2022. pp. 3268--3275 (2022)

\bibitem{wordalignment}
Ho, A.K.N., Yvon, F.: Optimizing word alignments with better subword tokenization. In: The 18th biennial conference of the International Association of Machine Translation (2021)

\bibitem{Hooker2018-vg}
Hooker, S., Erhan, D., Kindermans, P.J., Kim, B.: A benchmark for interpretability methods in deep neural networks. Advances in neural information processing systems  \textbf{32} (2019)

\bibitem{Hsieh_undated-vh}
Hsieh, C.Y., Yeh, C.K., Liu, X., Ravikumar, P., Kim, S., Kumar, S., Hsieh, C.J.: Evaluations and methods for explanation through robustness analysis. In: International Conference on Learning Representations (ICLR) (2020)

\bibitem{hussein2018homogeneous}
Hussein, N.O., Elttayef, A.I.: The homogeneous relationship between syntax and semantics: Structure, and meaning in arrangement of english language sentences. British Journal of English Linguistics  \textbf{6}(5),  38--45 (2018)

\bibitem{jacovi2020towards}
Jacovi, A., Goldberg, Y.: Towards faithfully interpretable nlp systems: How should we define and evaluate faithfulness? arXiv preprint arXiv:2004.03685  (2020)

\bibitem{jacovi-goldberg-2020-towards}
Jacovi, A., Goldberg, Y.: Towards faithfully interpretable {NLP} systems: How should we define and evaluate faithfulness? In: Jurafsky, D., Chai, J., Schluter, N., Tetreault, J. (eds.) Proceedings of the 58th Annual Meeting of the Association for Computational Linguistics. pp. 4198--4205. Association for Computational Linguistics, Online (Jul 2020). \doi{10.18653/v1/2020.acl-main.386}, \url{https://aclanthology.org/2020.acl-main.386}

\bibitem{explanationprediction}
Karimi, A.H., Muandet, K., Kornblith, S., Sch{\"o}lkopf, B., Kim, B.: On the relationship between explanation and prediction: A causal view. arXiv preprint arXiv:2212.06925  (2022)

\bibitem{TransShap}
Kokalj, E., {\v{S}}krlj, B., Lavra{\v{c}}, N., Pollak, S., Robnik-{\v{S}}ikonja, M.: {BERT} meets shapley: Extending {SHAP} explanations to transformer-based classifiers. In: Toivonen, H., Boggia, M. (eds.) Proceedings of the EACL Hackashop on News Media Content Analysis and Automated Report Generation. pp. 16--21. Association for Computational Linguistics, Online (Apr 2021), \url{https://aclanthology.org/2021.hackashop-1.3}

\bibitem{sentencepiece}
Kudo, T., Richardson, J.: Sentencepiece: A simple and language independent subword tokenizer and detokenizer for neural text processing. arXiv preprint arXiv:1808.06226  (2018)

\bibitem{lai2019human}
Lai, V., Tan, C.: On human predictions with explanations and predictions of machine learning models: A case study on deception detection. In: Proceedings of the conference on fairness, accountability, and transparency. pp. 29--38 (2019)

\bibitem{disentanglesynsem}
Li, D., Fei, H., Ren, S., Li, P.: A deep decomposable model for disentangling syntax and semantics in sentence representation. In: Findings of the Association for Computational Linguistics: EMNLP 2021. pp. 4300--4310 (2021)

\bibitem{OOD-GNN}
Li, H., Wang, X., Zhang, Z., Zhu, W.: {OOD-GNN}: {Out-of-Distribution} generalized graph neural network  (Dec 2021)

\bibitem{perturbscore}
Li, L., Ren, K., Shao, Y., Wang, P., Qiu, X.: Perturbscore: Connecting discrete and continuous perturbations in nlp. arXiv preprint arXiv:2310.08889  (2023)

\bibitem{perturbrobustness}
Li, L., Ren, K., Shao, Y., Wang, P., Qiu, X.: Perturbscore: Connecting discrete and continuous perturbations in nlp. arXiv preprint arXiv:2310.08889  (2023)

\bibitem{posthocxAI}
Madsen, A., Reddy, S., Chandar, S.: Post-hoc interpretability for neural nlp: A survey. ACM Computing Surveys  \textbf{55}(8),  1--42 (2022)

\bibitem{chunkalignment}
Majumder, G., Pakray, P., Das, R., Pinto, D.: Interpretable semantic textual similarity of sentences using alignment of chunks with classification and regression. Applied Intelligence  \textbf{51},  7322--7349 (2021)

\bibitem{Moradi2021EvaluatingTR}
Moradi, M., Samwald, M.: Evaluating the robustness of neural language models to input perturbations. In: Conference on Empirical Methods in Natural Language Processing (2021), \url{https://api.semanticscholar.org/CorpusID:237346776}

\bibitem{perturbationsurvey}
Moradi, M., Samwald, M.: Evaluating the robustness of neural language models to input perturbations. arXiv preprint arXiv:2108.12237  (2021)

\bibitem{nagahisarchoghaei2023empirical}
Nagahisarchoghaei, M., Nur, N., Cummins, L., Nur, N., Karimi, M.M., Nandanwar, S., Bhattacharyya, S., Rahimi, S.: An empirical survey on explainable ai technologies: Recent trends, use-cases, and categories from technical and application perspectives. Electronics  \textbf{12}(5), ~1092 (2023). \doi{10.3390/electronics12051092}

\bibitem{niu-etal-2020-evaluating}
Niu, X., Mathur, P., Dinu, G., Al-Onaizan, Y.: Evaluating robustness to input perturbations for neural machine translation. In: Jurafsky, D., Chai, J., Schluter, N., Tetreault, J. (eds.) Proceedings of the 58th Annual Meeting of the Association for Computational Linguistics. pp. 8538--8544. Association for Computational Linguistics, Online (Jul 2020). \doi{10.18653/v1/2020.acl-main.755}, \url{https://aclanthology.org/2020.acl-main.755}

\bibitem{prabhakaran-etal-2019-perturbation}
Prabhakaran, V., Hutchinson, B., Mitchell, M.: Perturbation sensitivity analysis to detect unintended model biases. In: Inui, K., Jiang, J., Ng, V., Wan, X. (eds.) Proceedings of the 2019 Conference on Empirical Methods in Natural Language Processing and the 9th International Joint Conference on Natural Language Processing (EMNLP-IJCNLP). pp. 5740--5745. Association for Computational Linguistics, Hong Kong, China (Nov 2019). \doi{10.18653/v1/D19-1578}, \url{https://aclanthology.org/D19-1578}

\bibitem{LIME}
Ribeiro, M.T., Singh, S., Guestrin, C.: " why should i trust you?" explaining the predictions of any classifier. In: Proceedings of the 22nd ACM SIGKDD international conference on knowledge discovery and data mining. pp. 1135--1144 (2016)

\bibitem{contrastivexai}
Ross, A., Marasovi{\'c}, A., Peters, M.E.: Explaining nlp models via minimal contrastive editing (mice). arXiv preprint arXiv:2012.13985  (2020)

\bibitem{Sajjad2022survey}
Sajjad, H., Durrani, N., Dalvi, F.: {Neuron-level Interpretation of Deep NLP Models: A Survey}. Trans. of the Association for Computational Linguistics  \textbf{10},  1285--1303 (11 2022). \doi{10.1162/tacl_a_00519}, \url{https://doi.org/10.1162/tacl\_a\_00519}

\bibitem{sennrich-etal-2016-neural}
Sennrich, R., Haddow, B., Birch, A.: Neural machine translation of rare words with subword units. In: Proc. of the 54th Annual Meeting of the Association for Computational Linguistics (Volume 1: Long Papers). pp. 1715--1725. Association for Computational Linguistics, Berlin, Germany (Aug 2016). \doi{10.18653/v1/P16-1162}, \url{https://aclanthology.org/P16-1162}

\bibitem{lmfingerprints}
Sevastjanova, R., Kalouli, A., Beck, C., Hauptmann, H., El-Assady, M.: Lmfingerprints: Visual explanations of language model embedding spaces through layerwise contextualization scores. In: Computer Graphics Forum. vol.~41, pp. 295--307. Wiley Online Library (2022)

\bibitem{ShapleyValue}
Shapley, L.S., et~al.: A value for n-person games. Princeton University Press Princeton (1953)

\bibitem{syntaxperturbation}
Sinha, S., Chen, H., Sekhon, A., Ji, Y., Qi, Y.: Perturbing inputs for fragile interpretations in deep natural language processing. arXiv preprint arXiv:2108.04990  (2021)

\bibitem{spinner2020explainer}
Spinner, T., Schlegel, U., Sch{\"{a}}fer, H., El{-}Assady, M.: explainer: A visual analytics framework for interactive and explainable machine learning. IEEE Trans. on Visualization and Computer Graphics  \textbf{26}(1),  1064--1074 (2020). \doi{10.1109/TVCG.2019.2934629}

\bibitem{OOD-pb}
Stalder, S., Perraudin, N., Achanta, R., Perez-Cruz, F., Volpi, M.: What you see is what you classify: Black box attributions. Advances in Neural Information Processing Systems  \textbf{35},  84--94 (2022)

\bibitem{truong-etal-2023-language}
Truong, T.H., Baldwin, T., Verspoor, K., Cohn, T.: Language models are not naysayers: an analysis of language models on negation benchmarks. In: Palmer, A., Camacho-collados, J. (eds.) Proceedings of the 12th Joint Conference on Lexical and Computational Semantics (*SEM 2023). pp. 101--114. Association for Computational Linguistics, Toronto, Canada (Jul 2023). \doi{10.18653/v1/2023.starsem-1.10}, \url{https://aclanthology.org/2023.starsem-1.10}

\bibitem{vijayakumar2023interpretability}
Vijayakumar, S.: Interpretability in activation space analysis of transformers: A focused survey. Proc. of the ACM Int. Conf. on Information and Knowledge Management Workshops  (2022)

\bibitem{necessitysufficiency}
Watson, D.S., Gultchin, L., Taly, A., Floridi, L.: Local explanations via necessity and sufficiency: Unifying theory and practice. In: Uncertainty in Artificial Intelligence. pp. 1382--1392. PMLR (2021)

\bibitem{goodenough}
Wilson, V.J., McCance, T.: Good enough evaluation  (2015)

\bibitem{wu-etal-2021-polyjuice}
Wu, T., Ribeiro, M.T., Heer, J., Weld, D.: Polyjuice: Generating counterfactuals for explaining, evaluating, and improving models. In: Zong, C., Xia, F., Li, W., Navigli, R. (eds.) Proceedings of the 59th Annual Meeting of the Association for Computational Linguistics and the 11th International Joint Conference on Natural Language Processing (Volume 1: Long Papers). pp. 6707--6723. Association for Computational Linguistics, Online (Aug 2021). \doi{10.18653/v1/2021.acl-long.523}, \url{https://aclanthology.org/2021.acl-long.523}

\bibitem{LMstochasticity}
Youvan, D.: Stochastic elements in language models: Exploring variability and randomness in ai responses  (11 2023). \doi{10.13140/RG.2.2.27309.26089}

\bibitem{zhao2024survey}
Zhao, H., Chen, H., Yang, F., Liu, N., Deng, H., Cai, H., Wang, S., Yin, D., Du, M.: Explainability for large language models: A survey. ACM Trans. Intell. Syst. Technol.  \textbf{15}(2) (feb 2024). \doi{10.1145/3639372}, \url{https://doi.org/10.1145/3639372}

\bibitem{zini2022survey}
Zini, J.E., Awad, M.: On the explainability of natural language processing deep models. ACM Computing Surveys  \textbf{55}(5) (dec 2022). \doi{10.1145/3529755}, \url{https://doi.org/10.1145/3529755}

\end{thebibliography}

\end{document}